\documentclass[runningheads]{llncs}

\usepackage[T1]{fontenc}
\usepackage{graphicx}
\usepackage{hyperref}
\hypersetup{
    colorlinks=true,
    linkcolor=black,
    citecolor=black,
    urlcolor=blue
}
\usepackage{glossaries}
\usepackage{subcaption}
\DeclareCaptionFormat{bold-figure}{\textbf{#1#2}#3\par}
\DeclareCaptionFormat{bold-table}{\textbf{#1#2}#3\par}
\usepackage{nicematrix}
\usepackage[numbers,sort&compress]{natbib}

\usepackage{xcolor,colortbl}

\begin{document}

\newacronym{os}{OS}{Operating System}
\newacronym{nyi}{NYI}{Not Yet Implemented}
\newacronym{dei}{DEI}{Department of Informatics Engineering}

\newacronym{EC}{EC}{Evolutionary Computation}
\newacronym{EA}{EA}{Evolutionary Algorithm}
\newacronym{GA}{GA}{Genetic Algorithm}
\newacronym{GP}{GP}{Genetic Programming}
\newacronym{GE}{GE}{Grammatical Evolution}
\newacronym{SGE}{SGE}{Structured Grammatical Evolution}
\newacronym{AutoML}{AutoML}{Automated Machine Learning}
\newacronym{AST}{AST}{Abstract Syntax Tree}
\newacronym{HL}{HL}{Hearing Loss}
\newacronym{FE}{FE}{Feature Engineering}
\newacronym{WHO}{WHO}{World Health Organization}
\newacronym{AI}{AI}{Artificial Intelligence}
\newacronym{ML}{ML}{Machine Learning}
\newacronym{CISUC}{CISUC}{Center for Informatics and Systems of the University of Coimbra}
\newacronym{GBGP}{GBGP}{Grammar-Based Genetic Programming}
\newacronym{CFG}{CFG}{Context-Free Grammar}
\newacronym{BNF}{BNF}{Backus-Naur Form}
\newacronym{HPO}{HPO}{Hyper-parameter Optimisation}
\newacronym{FC}{FC}{Feature Construction}
\newacronym{FS}{FS}{Feature Selection}
\newacronym{FMS}{FMS}{Full Model Selection}
\newacronym{CASH}{CASH}{Combined Algorithm Selection and Hyperparameter optimization problem}
\newacronym{CNN}{CNN}{Convolutional Neural Network}
\newacronym{ANN}{ANN}{Artificial Neural Network}
\newacronym{HyTEA}{HyTEA}{Hybrid Tree Evolutionary Algorithm}
\newacronym{DE}{DE}{Differential Evolution}
\newacronym{DT}{DT}{Decision Tree}
\newacronym{A4A}{A4A}{Audiology for All}
\newacronym{RF}{RF}{Random Forest}
\newacronym{ETL}{ETL}{Extract, Transform and Load}
\newacronym{ER}{ER}{Entity-Relationship}
\newacronym{ANOVA}{ANOVA}{Analysis of Variance}
\newacronym{PSGE}{PSGE}{Probabilistic Structured Grammatical Evolution}
\newacronym{NaN}{NaN}{Not a Number}
\newacronym{PCA}{PCA}{Principal Component Analysis}
\newacronym{SOM}{SOM}{Self-Organizing Map}
\newacronym{AE}{AE}{Autoencoder}
\newacronym{UMAP}{UMAP}{Uniform Manifold Approximation and Projection}
\newacronym{XGB}{XGB}{Extreme Gradient Boosting}
\newacronym{MLP}{MLP}{Multi-Layer Perceptron}

\newcommand{\plotsummary}[2]{    
    \begin{figure}
        \centering
        \begin{subfigure}[b]{0.49\textwidth}
            \centering
            \includegraphics[page=2,width=\textwidth]{#1}
            \caption{Fitness Plot}
            \label{fig:#2:fitness}
        \end{subfigure}
        \begin{subfigure}[b]{0.49\textwidth}
            \centering
            \includegraphics[page=3,width=\textwidth]{#1}
            \caption{Feature Evolution Plot}
            \label{fig:#2:feature-evolution}
        \end{subfigure}
        \begin{subfigure}[b]{0.49\textwidth}
            \centering
            \includegraphics[page=4,width=\textwidth]{#1}
            \caption{Feature Complexity}
            \label{fig:#2:ratios}
        \end{subfigure}
        \begin{subfigure}[b]{0.49\textwidth}
            \centering
            \includegraphics[page=5,width=\textwidth]{#1}
            \caption{Number of Features}
            \label{fig:#2:distribution}
        \end{subfigure}
        \begin{subfigure}[b]{0.96\textwidth}
            \centering
            \includegraphics[page=6,width=\textwidth]{#1}
            \caption{Feature Engineering Methods Comparison}
            \label{fig:#2:comparison}
        \end{subfigure}
        \caption{Experiment Summary - #2}
        \label{fig:#2}
    \end{figure}
}

\title{Decision Tree Based Wrappers for Hearing Loss}
%
%
\author{Miguel Rabuge\orcidID{0009-0008-0914-0495} \and
Nuno Lourenço\orcidID{0000-0002-2154-0642}}
\authorrunning{M. Rabuge and N. Lourenço}
%
\institute{
CISUC/LASI – Centre for Informatics and Systems of the University of Coimbra, Department of Informatics Engineering, University of Coimbra
    \email{\{rabuge,naml\}@dei.uc.pt}\\
}
\maketitle              
\begin{abstract}
Audiology entities are using \gls{ML} models to guide their screening towards people at risk. \gls{FE} focuses on optimizing data for \gls{ML} models, with evolutionary methods being effective in feature selection and construction tasks. This work aims to benchmark an evolutionary \gls{FE} wrapper, using models based on decision trees as proxies. The FEDORA framework is applied to a \gls{HL} dataset, being able to reduce data dimensionality and statistically maintain baseline performance. Compared to traditional methods, FEDORA demonstrates superior performance, with a maximum balanced accuracy of 76.2\%, using 57 features. The framework also generated an individual that achieved 72.8\% balanced accuracy using a single feature.

\keywords{Feature Engineering \and Grammatical Evolution \and Audiology}
\end{abstract}
\section{Introduction}
\glsresetall
The advances in our digital world have brought us large amounts of data that can be used to extract domain-specific knowledge. One such domain is the medical field, where data is used to help professionals better decide through its analysis, visualization and usage in decision support systems.

The medical field encompasses a broad range of disciplines, including audiology. As a specialized branch within the medical field, it focuses on studying hearing, balance, and associated disorders. In February 2024, the \gls{WHO} reiterates its prediction that by 2050, 2.5 billion people will have \gls{HL}, with 1 in 10 requiring rehabilitation \cite{whohl}. This condition can negatively impact a person's life, either professionally or personally.

As such, audiology technicians are conducting screenings to assess the hearing health of the population, while collecting data that can help guide the screening towards people at risk, through intelligent models.

This can be achieved by \gls{ML} models that provide a wide range of methods to detect and predict patterns. One key aspect of properly modelling them is defining the data representation that is given as input. \gls{FE} is a step in the \gls{ML} pipeline dedicated to transforming data to suit the requirements of these models. Despite existing methods to address this problem, evolutionary methods have demonstrated their utility for selecting and constructing novel features.

This work aims to benchmark an evolutionary \gls{FE} wrapper, using models based on decision trees as proxies. The FEDORA framework will be applied to a \gls{HL} classification dataset, in three different settings, varying only on the choice of the proxy, which can be a \gls{DT} or its bagging and boosting variants: \gls{RF} and \gls{XGB}, respectively.

Results confirm that FEDORA can reduce the dimensionality of the data while statistically maintaining baseline performance, in every experiment. The framework is compared with common \gls{FE} methods and consistently outperforms them, with statistical significance and large effect sizes. The best result obtained is 76.2\% balanced accuracy using an individual from the \gls{RF} proxy experiment, and a \gls{XGB} as the testing model, using 57 features that were selected or constructed from the 60 original ones. When using the least amount of features, the best result is 72.8\% balanced accuracy using an individual from the \gls{DT} proxy experiment and a \gls{RF} algorithm as the testing model, using a single feature.

\section{Related Work}
\subsection{Evolutionary Feature Engineering}

As a step of the \gls{ML} pipeline, \gls{FE} defines the process of transforming an original dataset into a refined one. It can be partitioned into two domains: \gls{FS} and \gls{FC}. The goal of \gls{FS} is to remove redundant or misleading features that can compromise the performance of the models. In addition, \gls{FC} seeks to build new features from the original ones, providing an enhanced representation that may help \gls{ML} models, especially those that cannot create a complex internal representation or decision boundary.

There are three main types of \gls{FE} methods: filter, wrapper and embedded. \cite{cherrier2019consistent}. Filter methods assess the features without the use of a \gls{ML} model. In contrast, wrapper methods use the performance of such models to evaluate the set of features, which is the approach this work follows. At last, embedded methods perform \gls{FE} while training the model.

Evolutionary \gls{FE} methods have been proposed over the years with \gls{GP} \cite{koza1994genetic} being the most common approach. Concerning approaches that use \gls{DT}-based proxies, \citet{tran2016multiple} proposed MultGPFC, a hybrid (filter and wrapper) framework that uses a \gls{DT} proxy and a filter distance metric. The fitness function is given by a linear combination of both approaches, with the accuracy of the \gls{DT} being the average score of a 3-fold cross-validation repeated 3 times with different data splits. The framework was applied to 6 datasets, showing that it can construct and select features that boost the performance of \gls{ML} testing models, although being more effective for a \gls{DT}. \citet{cherrier2019consistent} also followed a \gls{GP} approach to design and compare evolutionary wrapper or filter methods that construct interpretable features for three experimental physics datasets. Among the methods, the 3-fold cross-validation accuracy of a \gls{DT} and \gls{XGB} models were used to evaluate the individuals, in different experiments. Whether evolving one or more features, all methods improved the baseline.

Regarding \gls{GE} \cite{ryan1998grammatical} works, \citet{miquilini2016enhancing} compared two types of \gls{DT} algorithms as proxies, namely J48 and REPTree, for evolving a single feature. The fitness of the individuals was measured in a 5-fold cross-validation setting and given by its average accuracy. Being applied to 16 datasets, both proxies produced features that empowered the corresponding models with higher performance and a smaller tree depth than the baseline, for most problems. Additionally, the work of \citet{monteiro2021fermat} proposed FERMAT, a framework that uses \gls{SGE} \cite{lourencco2016sge,lourencco2018structured}, a \gls{GE} variant, as the evolutionary engine. In this work, a \gls{DT} is used as the proxy for a \gls{RF}, the testing model. The fitness of the individuals was given by the validation Root Mean Squared Error (RMSE) of the proxy. It was applied to two regression problems, having success in selecting and constructing new features that helped regression models achieve better predictions.

\subsection{Machine Learning in Hearing Loss Detection}

The current status of \gls{HL} detection by \gls{ML} models is overviewed in the work of \citet{miranda2022hytea}. Most works focus on actively detecting \gls{HL} through the results of audiology screenings or related procedures, demographics, medical data and noise exposure metrics. These features generally match with the ones highlighted by the \gls{WHO} as relevant \gls{HL} causes. Frequently, studies on \gls{HL} detection focus on specific categories or origins of \gls{HL}, such as sensorineural or noise-induced causes, as well as environments where \gls{HL} is prevalent, such as industrial settings \cite{tomiazzi2019performance}. 

Results show that with screening or similar information, \gls{ML} models can achieve accuracy values above 70\%, depending on the data and model used. However, when aiming to guide screening towards people at risk, it is expensive to perform a screening procedure across the whole population. Therefore, models that rely solely on personal, medical, and demographic factors to predict the likelihood of \gls{HL}, in the absence of screening data, could be valuable for discerning which contextual factors have a greater impact on \gls{HL}.

\section{Approach}

\begin{figure}
    \centering
    \includegraphics[scale=0.5]{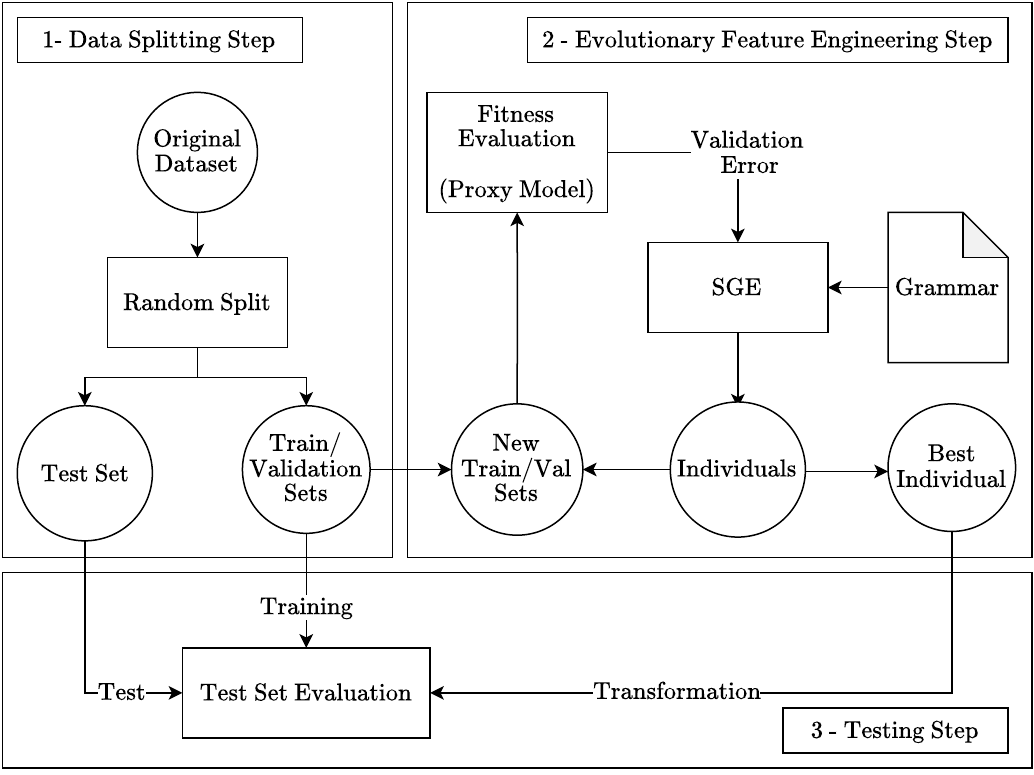}
    \caption{FEDORA: Feature Engineering through Discovery of Reliable Attributes}
    \label{fig:fedora}
\end{figure}

In this work, FEDORA \cite{rabuge2024comparison} will be applied to a \gls{HL} detection problem using three distinct classifiers based on decision trees, namely basic \gls{DT} models and their bagging and boosting counterparts, \gls{RF} and \gls{XGB}, as proxy models in the evolutionary framework. Figure \ref{fig:fedora} illustrates the inner workings of the framework.

The framework starts by splitting the original dataset into training (40\%), validation (40\%) and test subsets (20\%). The training and validation subsets are given to the evolutionary process, where \gls{SGE} will generate individuals that select and construct a new dataset from the original one, through a context-free grammar. These transformations will be applied to the training and validation subsets, which will then be used to train the proxy model and validate the transformation, respectively. The fitness is given by the validation error, namely (1 - Balanced Accuracy). After the specified generations of the evolutionary process, the individual with the lowest validation error is returned. This individual is then applied to the three subsets and its ability to generalize to unseen data is evaluated. This assessment involves training a range of Machine Learning (ML) models using both the training and validation subsets and subsequently evaluating their performance on the test set.
\section{Experimental Setup}

This study addresses detecting \gls{HL} with contextual attributes through binary classification. The dataset generation process is fully defined in \cite{miranda2022hytea}. The dataset has 60 features and cannot be publicly published due to sensitive patient screening information.

Regarding the experimental settings, Table \ref{table:setup} summarizes the parameters of the framework for each one of the three experiments. Most settings are alike, only diverging in the proxy model. All the models used the default package parameters, except for the \gls{RF} where the n\_estimators and max\_depth parameters were defined to 5. The grammar used in the experiments enables the selection and construction of algebraic-type features and is available here \footnote{\href{https://github.com/miguelrabuge/fedora/blob/main/examples/audiology/audiology.pybnf}{ github.com/miguelrabuge/fedora/blob/main/examples/audiology/audiology.pybnf}}.

\begin{table}
\centering
\caption{Experimental Settings}
\label{table:setup}
\begin{NiceTabular}{|l|wc{2cm}|wc{2cm}|wc{2cm}|}
\hline
\multicolumn{1}{|c}{\textbf{Parameters}} & \multicolumn{3}{c|}{\textbf{Experiments}} \\
\hline
Proxy Model & DT & RF & XGB \\
\hline
Population & \multicolumn{3}{c|}{200} \\
\hline
Generations & \multicolumn{3}{c|}{100} \\
\hline
Runs & \multicolumn{3}{c|}{30} \\
\hline
Elitism & \multicolumn{3}{c|}{10\%} \\
\hline
Crossover Rate & \multicolumn{3}{c|}{0.9} \\
\hline
Mutation Rate & \multicolumn{3}{c|}{0.1} \\
\hline
Minimum Tree Depth & \multicolumn{3}{c|}{3} \\
\hline
Maximum Tree Depth & \multicolumn{3}{c|}{10} \\
\hline
Selection & \multicolumn{3}{c|}{Tournament (size 3)} \\
\hline
Fitness & \multicolumn{3}{c|}{1 - Balanced Accuracy} \\
\hline
\end{NiceTabular}
\end{table}

Four types of models were selected as testing models: \gls{DT}, \gls{RF}, \gls{XGB} and \gls{MLP}. These models will assess the generalization performance of the FEDORA individuals, comparing its balanced accuracy scores with the baseline and other \gls{FE} methods, such as \gls{PCA}, \gls{UMAP}, \glspl{SOM} and \glspl{AE}.

Each \gls{FE} technique will use the same number of features as the FEDORA individual. For instance, if the FEDORA individual has 15 features, both the number of \gls{PCA} and \gls{UMAP} components would be equal to 15, the 2D SOM grid would have dimensions of 15x1, and the code size of the AE would be set to 15. The \gls{AE} parameters consist of 50 neurons for the single hidden layers, with linear activation functions, and using mean squared error as the error metric. Its training involves using a batch size of 32, running for 50 epochs, using Stochastic Gradient Descent.
\section{Results and Discussion}

The evolution process will be examined from the perspectives of fitness and the number of features, considering the average values across 30 runs over the generations. This allows us to overview the evolution process from both perspectives, checking for relevant behaviours. The best individuals will be analysed via the number and construction complexity of the features they generate. This gives us insights from both \gls{FS} and \gls{FC} standpoints. The performance results of \gls{ML} classifiers, using various \gls{FE} methods, will also be visually examined and statistically analysed to check for meaningful differences.

\subsection{Using Decision Trees as Proxy}

Figure \ref{fig:DT} showcases a collection of plots depicting the results of the \gls{DT}  experiment from different perspectives. In Panel \ref{fig:DT:fitness}, the evolution of the average fitness of populations and the performance of the best individuals across 30 runs is depicted over successive generations, showing an effective minimization trend of the balanced accuracy validation error. The population line reaches an average error mark of 32\%, while the best line achieves a lower error of 29\%.

In Panel \ref{fig:DT:feature-evolution}, four distinct lines are displayed, each representing the average number of features selected by FEDORA across 30 runs. These lines correspond to the averages of the population (population), the best individual (best), and individuals with the least (minimum) and greatest (maximum) number of features. The minimum and maximum lines are roughly around both ends of the number of allowed features by the grammar. Conversely, the best and population lines have been decreasing over the generations, without any signs of stabilizing, despite having an initial increase. These two panels show that using a \gls{DT} as the proxy model induces the framework to maximise performance and reduce the number of features over the generations.

Panel \ref{fig:DT:ratios} illustrates feature ratios derived from the best individual of each run. To construct this chart, we establish criteria for classifying features produced by FEDORA individuals. A feature is named as \textit{original} if it is solely selected from the original dataset (e.g. feature1), \textit{engineered} if a single operator merges two original features (e.g. feature1 + feature2), and \textit{complex} if two or more operators are utilized (e.g. feature1 + feature2 - feature3). Also, Panel \ref{fig:DT:distribution} must be considered when interpreting this one, as it provides the total number of features for each best individual. The feature complexity ratios are normalized by these values, as shown in the equations below.
\begin{equation*}
R_{O} = \frac{N_{Selected}}{N_{Total}} \hspace{1cm} R_{E} = \frac{N_{Engineered}}{N_{Total}}\hspace{1cm} R_{C} = \frac{N_{Complex}}{N_{Total}}
\end{equation*}

Therefore, Panel \ref{fig:DT:ratios} shows that the individuals are composed of constructed and selected features since the ratio of original features and the sum of engineered and complex features ratios are both positive. Some individuals present large ratios of engineered and complex figures due to having a low number of features, as observed in Panel \ref{fig:DT:distribution}. Runs 6, 8 and 25 returned individuals without original features, only being composed of engineered or complex features. 

To compare FEDORA with the baseline and other common \gls{FE} methods, Panel \ref{fig:DT:comparison} exhibits a series of 24 boxplots associated with the testing outcomes. Each boxplot contains 30 points, representing each run individually. The value of each point corresponds to the balanced accuracy score of the respective \gls{FE} method and testing model pipeline in a particular run. When using a \gls{DT} as the testing model, the FEDORA boxplot visually improves baseline performance, while slightly deteriorating it on the other \gls{ML} models. Examining the remaining \gls{FE} techniques, most underperform the baseline and FEDORA in all testing models. When aiming for maximum performance with minimal features, run 8 returned an individual that achieves a 72.8\% balanced accuracy score with a \gls{RF} classifier, with a single complex feature. Its phenotype is shown below:
$$x_{29}-x_{22}*x_{22}+(x_{8}*x_{42}/(x_{35}*x_{9}+x_{53}))$$

\plotsummary{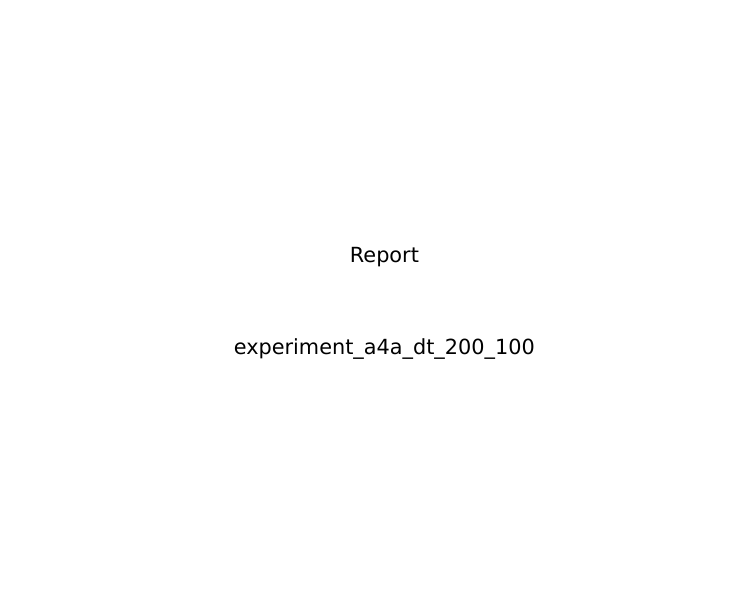}{DT}

\subsection{Using Random Forests as Proxy}

Figure \ref{fig:RF} presents the same set of plots with the results of the RF experiment. In Panel \ref{fig:RF:fitness}, one can see that both lines stabilize around the 20-generation mark, with slight improvements observed in the subsequent generations, resulting in a final average validation error of 29\% for the population line and 26\% for the best line.

Panel \ref{fig:RF:feature-evolution} shows that the population and best lines exhibit simultaneous growth up to the point of reaching 50 features, with the latter slightly surpassing the former, a behaviour that contrasts with the \gls{DT} experiment. Also, the minimum and maximum lines are close to the 0 and 60 features mark, respectively.

Panel \ref{fig:RF:ratios} demonstrates that the features generated by the individuals are approximately 80\% original, 10\% engineered and 10\% complex, while mostly using less than 60 features, as shown in Panel \ref{fig:RF:distribution}. Although differently than observed in the \gls{DT} experiment, these Panels reinforce the claim that the framework can simultaneously select and construct features.

Regarding the testing results, Panel \ref{fig:RF:comparison} shows that FEDORA maintains baseline performance, despite using fewer features. It also outperforms the remaining \gls{FE} methods, which deteriorate baseline performance across all testing classifiers. The best-performing individual was obtained in run 19, with a 76.2\% balanced accuracy score, using the \gls{XGB} classifier with 57 total features (45 Original, 6 Engineered and 6 Complex). It corresponds to the best score obtained in this paper by the proposed framework.

\plotsummary{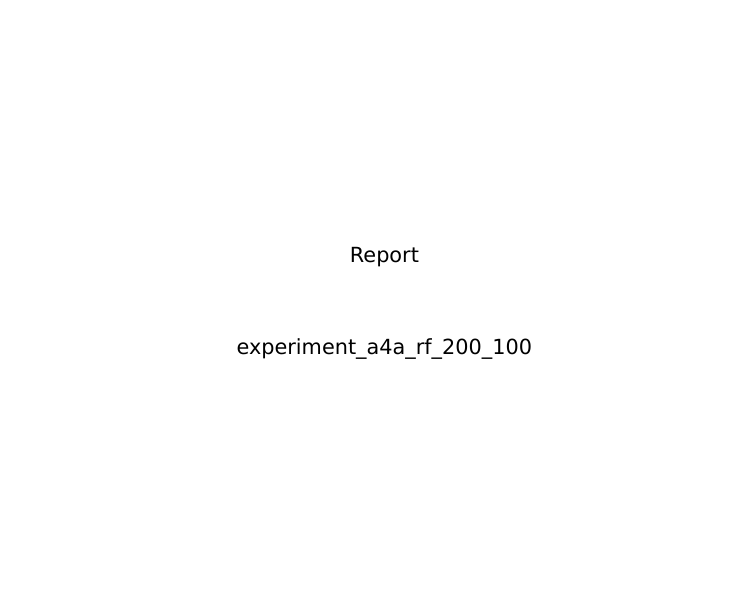}{RF}

\subsection{Using Extreme Gradient Boosting as Proxy}

Figure \ref{fig:XGB} summarizes the obtained results of the \gls{XGB} experiment. Similarly to the RF experiment, Panel \ref{fig:XGB:fitness} shows a clear effective minimization of the error, this time achieving lower error scores with both lines stabilizing earlier, at the 10-generation mark, with the population line achieving an error of around 27\% and the best an error of roughly 25\%.

The analysis made for the Panels \ref{fig:RF:feature-evolution}, \ref{fig:RF:ratios} and \ref{fig:RF:distribution} of the RF experiment is directly applicable to the Panels \ref{fig:XGB:feature-evolution}, \ref{fig:XGB:ratios} and \ref{fig:XGB:distribution} of this experiment, i.e. FEDORA can perform \gls{FS} and \gls{FC} since the original ratio and the sum of the remaining ratios are positive, correspondingly. Although returning individuals with a slightly greater amount of features, the evolution and complexity ratios of the features in this experiment are similar to the \gls{RF} proxy experiment.

In Panel \ref{fig:XGB:comparison}, FEDORA can maintain baseline performance across all classifiers. The framework outperforms common \gls{FE} methods, especially when using the \gls{RF} and \gls{XGB} classifiers. It is possible to observe a narrow improvement over the baseline with the \gls{XGB} classifier when using the FEDORA individuals. The best-performing individual of this experiment was obtained in run 19, with a 76\% balanced accuracy score, using the \gls{XGB} classifier with 58 total features (39 Original, 13 Engineered and 6 Complex).

\plotsummary{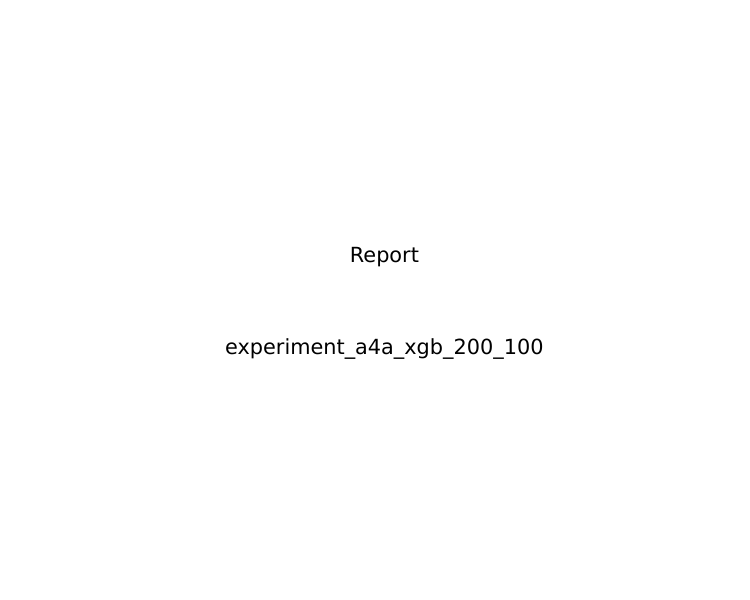}{XGB}

\subsection{Statistical Analysis}
To compare the results of the different experiments, we performed a statistical analysis to check for any meaningful differences. The statistical tests were only applied to the \gls{FE} methods of one single testing classifier for each experiment, for simplicity. The chosen testing model for the statistical test is the same as the proxy of the corresponding experiment.

\setlength{\tabcolsep}{0.5em} 
{\renewcommand{\arraystretch}{1}
Without making any parametric or paired assumptions, the Kruskal-Wallis non-parametric test was applied to compare the \gls{FE} techniques, in each experiment, to check if the median scores of all the groups are equal, with a significance level of 0.05. Table \ref{table:kruskal} gives the Kruskal-Wallis test results for every experiment. As the p-value is 0 for all experiments, every experiment rejects the null hypothesis, i.e. there are differences in the medians of the groups. Therefore, a pairwise post hoc analysis is required for every pair of groups in each experiment. 

Pair-wise comparisons were made using Dunn's posthoc test and correcting the resulting p-values with the Bonferroni correction. Cliff's $\delta$ was used to measure the effect size. The symbol "$\sim$" denotes a negligible effect size ($|\delta| < 0.147$), "+" denotes a small effect size ($0.147 \le |\delta| < 0.33$), "++" a medium one ($0.33 \le |\delta| < 0.474$) and "+++" a large one ($|\delta| \ge 0.474$).

Table \ref{table:dunn-dt_200_100} details the effect sizes for Dunn's posthoc analysis for the \gls{DT} proxy experiment. It shows statistically significant differences between FEDORA and the other \gls{FE} methods, with a large effect size. There are also differences between the baseline and the common \gls{FE} methods, with a large effect size. For this experiment, there is no evidence of differences between the baseline and the FEDORA groups, meaning that the framework can statistically maintain performance. There are statistically significant differences between the \gls{UMAP} and the \gls{ANN} based \gls{FE} methods, i.e. the \glspl{SOM} and the \glspl{AE}, both with large effect sizes.

Table \ref{table:dunn-rf_200_100} provides the effect sizes for the \gls{RF} proxy experiment. Once again, the baseline and the proposed framework have statistically significant differences with the common \gls{FE} methods. Also, the baseline and FEDORA groups do not seem to have differences. Furthermore, there are statistically significant differences between the \gls{PCA} and \gls{UMAP} groups and between the \gls{AE} and \gls{UMAP} groups, with large effect sizes. Table \ref{table:dunn-xgb_200_100} gives the effect sizes for the \gls{XGB} experiment. The statistical analysis is the same as the one made for Table \ref{table:dunn-rf_200_100} since the tables are identical.

\begin{table}[h!]
    \centering
    \caption{Kruskal-Wallis Test Results}
    \label{table:kruskal}
    \begin{tabular}{|c|c|c|c|}
        \hline
         Experiment    & Model                  &   H &   P-Value \\
        \hline
         DT  & DecisionTreeClassifier &          143.45 &         0 \\
         RF  & RandomForestClassifier &          156.48 &         0 \\
         XGB & XGBClassifier          &          145.27 &         0 \\
        \hline
    \end{tabular}
\end{table}

\begin{table}[h!]
\centering
\caption{Dunn's test effect sizes - DT}
\label{table:dunn-dt_200_100}
    \begin{tabular}{|c|c|c|c|c|c|}
        \hline
         DT       & Baseline   & FEDORA           & PCA              & UMAP             & SOM              \\
        \hline
         FEDORA   &            & \cellcolor{gray} & \cellcolor{gray} & \cellcolor{gray} & \cellcolor{gray} \\
         PCA      & +++        & +++              & \cellcolor{gray} & \cellcolor{gray} & \cellcolor{gray} \\
         UMAP     & +++        & +++              &                  & \cellcolor{gray} & \cellcolor{gray} \\
         SOM      & +++        & +++              &                  & +++              & \cellcolor{gray} \\
         AE       & +++        & +++              &                  & +++              &                  \\
        \hline
    \end{tabular}
\end{table}

\begin{table}[h!]
\centering
\caption{Dunn's test effect sizes - RF}
\label{table:dunn-rf_200_100}
    \begin{tabular}{|c|c|c|c|c|c|}
        \hline
         RF             & Baseline         & FEDORA           & PCA              & UMAP             & SOM               \\
        \hline
         FEDORA         &                  & \cellcolor{gray} & \cellcolor{gray} & \cellcolor{gray} & \cellcolor{gray}  \\
         PCA            & +++              & +++              & \cellcolor{gray} & \cellcolor{gray} & \cellcolor{gray}  \\
         UMAP           & +++              & +++              & +++              & \cellcolor{gray} & \cellcolor{gray}  \\
         SOM            & +++              & +++              &                  &                  & \cellcolor{gray}  \\
         AE             & +++              & +++              &                  & +++              &                   \\
        \hline
    \end{tabular}
\end{table}

\begin{table}[h!]
\centering
\caption{Dunn's test effect sizes - XGB}
\label{table:dunn-xgb_200_100}
   \begin{tabular}{|c|c|c|c|c|c|}
       \hline
        XGB         & Baseline         & FEDORA           & PCA              & UMAP             & SOM               \\
       \hline
        FEDORA      &                  & \cellcolor{gray} & \cellcolor{gray} & \cellcolor{gray} & \cellcolor{gray}  \\
        PCA         & +++              & +++              & \cellcolor{gray} & \cellcolor{gray} & \cellcolor{gray}  \\
        UMAP        & +++              & +++              & +++              & \cellcolor{gray} & \cellcolor{gray}  \\
        SOM         & +++              & +++              &                  &                  & \cellcolor{gray}  \\
        AE          & +++              & +++              &                  & +++              &                   \\
       \hline
   \end{tabular}
\end{table}
}
\subsection{Discussion}
Concerning the evolution plots, all fitness plots show that individuals are gradually evolving throughout the generations. When using a \gls{DT} model as the proxy, the best line appears to be the one with greater evolution progress, although not quite matching the lower performances of the remaining experiments.

The feature evolution plots show a different angle of evolution. The number of features of the best individuals in the \gls{DT} experiment is decreasing throughout the generations, alongside the population mean. Such an event is not noticeable in the other experiments. The exact opposite happens, i.e. the best and population lines tend to grow and stabilize, with the latter resembling a logarithmic function. By observing the number of features in the \gls{DT} experiment, it is noticeable that its individuals can achieve a much lower feature dimensionality. This experiment also shows a higher ratio of engineered and complex features, although having fewer features biasing them. For the \gls{RF} and \gls{XGB} experiments, it is possible to observe that FEDORA can simultaneously select and construct novel features since the ratio of original features and the sum of engineered and complex features ratios are positive. 

Regarding the comparison with other common \gls{FE} methods and the baseline, the comparison plots show that FEDORA is consistently above the \gls{PCA}, \gls{UMAP}, \glspl{SOM} and \glspl{AE} methods while statistically maintaining baseline performance. In the \gls{DT} experiment, FEDORA is also able to improve past the baseline values when using a \gls{DT} as the testing model, although such results are not statistically significant.

From the analysed experiments, a pattern emerges in the behaviour of FEDORA. The \gls{DT} experiment can reduce the number of features to a degree that the other proxy models cannot. When comparing the inner workings of the proxy models, the \gls{RF} and \gls{XGB} models have one thing in common that the \gls{DT} model does not: the ability to create a more complex internal representation of the given data or decision boundary, which generally translates into better performances. A \gls{DT} can only make simple decisions with the provided data, which translates into axis-parallel hyper-planes decisions in the feature space, which might not properly address a complex dataset. As such, if the evolution transformations do not provide adequate features to this model, i.e. constructed features that allow for non-linear decisions in the original feature space, the \gls{DT} will most likely have worse performance than the remaining models, when facing a hard problem. Consequently, this encourages evolution to provide well-engineered features, thus making the fitness function much more discriminant. On the other hand, the remaining models do not put this kind of pressure on the evolution process. Each model takes charge of either constructing its features internally or defining a more complex decision boundary. Therefore, evolution just gives it a solid amount of original features, so that the model can find what works best for itself, and a few suggestions in the form of engineered and complex features. Consequently, the best individuals tend to have a much higher number of features when using \gls{RF}, or \gls{XGB} models as proxies. When using these models as the proxy, aiming for individuals with a low number of features becomes a problem. As such, ways to bias the evolution may be required, namely reducing the number of features that a transformation can produce in the grammar, e.g. 1 to 10 instead of 1 to 60, or adding a fitness component that penalizes individuals with many features. The usage of different feature combining operators may also be of use. These modifications might prove themselves useful in such a task.

Given these results and considering that FEDORA and the other methods usually work with fewer features, with their main purpose being a \gls{FE} technique, effectively reducing the number of features and statistically maintaining the baseline performance are great results. From the methods used in this work, FEDORA is the only one that can almost always have this behaviour. Also, it is possible to understand the phenotype of a FEDORA individual to a certain degree, depending on the choice of the operators defined in the grammar.

\section{Conclusion}

This work analysed the results of evolutionary wrapper approaches using decision tree based models as proxies and compared them with common \gls{FE} techniques on a \gls{HL} detection problem. Three experiments were conducted using the proposed framework, each employing different proxy models.

When comparing the three experiments, an interesting behaviour of the framework was discovered, when changing the proxy model. The \gls{DT} experiment drastically reduced the number of features, while the other models did not. To further reduce the number of features, one could bias the grammar or apply some penalty in the fitness function for the individuals that use a large number of features. This might not change the behaviour when using different models other than a \gls{DT}, but it forcefully reduces the number of features.  

The results confirm that FEDORA can reduce the dimensionality of the data while statistically maintaining baseline performance, in every experiment. The framework consistently outperforms the remaining \gls{FE} methods, with statistical significance and large effect sizes, proving itself as a viable alternative.

The best result obtained is 76.2\% balanced accuracy using an individual from the \gls{RF} experiment, and a \gls{XGB} algorithm as the testing model, using 57 total features (45 Original, 6 Engineered and 6 Complex) out of the 60 original ones. When using the least amount of features, the best result is 72,8\% balanced accuracy using an individual from the \gls{DT} experiment and a \gls{RF} algorithm as the testing model, using a single complex feature.

In future work, exploring the above-mentioned behaviours might be relevant to better understanding them, namely when biasing the grammar or penalizing the use of many features in the fitness function. Concerning the explainability of the FEDORA transformations, researching meaningful grammar operators might prove useful in addressing problem-specific needs. In this case, having logical operators for the boolean features, which have values of "yes" or "no", and the choice of a simple decision algorithm as the proxy, may increase explainability. Additionally, the previous study has identified several areas for future research, yet to be addressed. For instance, comparing the framework with other common and more complex methods and completing the full \gls{ML} pipeline through the use of a method that addresses the \gls{CASH}, such as \cite{assunccao2020evolution}, and comparing it to other full pipeline frameworks, could be beneficial for contextualizing and evaluating the framework within the \gls{AutoML} and \gls{EC} domains. The framework still needs to be analysed with different datasets to properly assess its generalization capabilities.

\begin{credits}
\subsubsection{\ackname} This work was partially funded by project A4A: Audiology for All (CENTRO-01-0247- FEDER-047083) financed by the Operational Program for Competitiveness and Internationalisation of PORTUGAL 2020 through the European Regional Development Fund, by project No. 7059 - Neuraspace - AI fights Space Debris, reference C644877546-00000020, supported by the RRP - Recovery and Resilience Plan and the European Next Generation EU Funds, following Notice No. 02/C05-i01/2022, Component 5 - Capitalization and Business Innovation - Mobilizing Agendas for Business Innovation, based upon work from COST Action Randomised Optimisation Algorithms Research Network (ROAR-NET), CA22137, supported by COST (European Cooperation in Science and Technology). This work is financed through national funds by FCT - Fundação para a Ciência e a Tecnologia, I.P., in the framework of the Project UIDB/00326/2020 and UIDP/00326/2020.

\subsubsection{\discintname}
The authors have no relevant competing interests to declare that are relevant to the content of this article.
\end{credits}
%
%

\bibliographystyle{splncs04}
\bibliography{references}

\begin{thebibliography}{13}
\providecommand{\natexlab}[1]{#1}
\providecommand{\url}[1]{\texttt{#1}}
\providecommand{\urlprefix}{URL }
\expandafter\ifx\csname urlstyle\endcsname\relax
  \providecommand{\doi}[1]{doi:\discretionary{}{}{}#1}\else
  \providecommand{\doi}{doi:\discretionary{}{}{}\begingroup \urlstyle{rm}\Url}\fi

\bibitem[{Assun{\c{c}}{\~a}o et~al.(2020)Assun{\c{c}}{\~a}o, Louren{\c{c}}o, Ribeiro, and Machado}]{assunccao2020evolution}
Assun{\c{c}}{\~a}o, F., Louren{\c{c}}o, N., Ribeiro, B., Machado, P.: Evolution of scikit-learn pipelines with dynamic structured grammatical evolution. In: International Conference on the Applications of Evolutionary Computation (Part of EvoStar), pp. 530--545, Springer (2020)

\bibitem[{Cherrier et~al.(2019)Cherrier, Poli, Defurne, and Sabati{\'e}}]{cherrier2019consistent}
Cherrier, N., Poli, J.P., Defurne, M., Sabati{\'e}, F.: Consistent feature construction with constrained genetic programming for experimental physics. In: 2019 IEEE Congress on Evolutionary Computation (CEC), pp. 1650--1658, IEEE (2019)

\bibitem[{Koza(1994)}]{koza1994genetic}
Koza, J.R.: Genetic programming as a means for programming computers by natural selection. Statistics and computing \textbf{4}, 87--112 (1994)

\bibitem[{Louren{\c{c}}o et~al.(2018)Louren{\c{c}}o, Assun{\c{c}}{\~a}o, Pereira, Costa, and Machado}]{lourencco2018structured}
Louren{\c{c}}o, N., Assun{\c{c}}{\~a}o, F., Pereira, F.B., Costa, E., Machado, P.: Structured grammatical evolution: a dynamic approach. Handbook of grammatical evolution pp. 137--161 (2018)

\bibitem[{Louren{\c{c}}o et~al.(2016)Louren{\c{c}}o, Pereira, and Costa}]{lourencco2016sge}
Louren{\c{c}}o, N., Pereira, F.B., Costa, E.: Sge: a structured representation for grammatical evolution. In: Artificial Evolution: 12th International Conference, Evolution Artificielle, EA 2015, Lyon, France, October 26-28, 2015. Revised Selected Papers 12, pp. 136--148, Springer (2016)

\bibitem[{Miquilini et~al.(2016)Miquilini, Barros, de~Melo, and Basgalupp}]{miquilini2016enhancing}
Miquilini, P., Barros, R.C., de~Melo, V.V., Basgalupp, M.P.: Enhancing discrimination power with genetic feature construction: A grammatical evolution approach. In: 2016 IEEE Congress on Evolutionary Computation (CEC), pp. 3824--3831, IEEE (2016)

\bibitem[{Miranda(2022)}]{miranda2022hytea}
Miranda, F.: HyTEA-Hybrid Tree Evolutionary Algorithm for Hearing Loss Diagnosis. Master's thesis (2022)

\bibitem[{Monteiro et~al.(2021)Monteiro, Louren{\c{c}}o, and Pereira}]{monteiro2021fermat}
Monteiro, M., Louren{\c{c}}o, N., Pereira, F.B.: Fermat: Feature engineering with grammatical evolution. In: EPIA Conference on Artificial Intelligence, pp. 239--251, Springer (2021)

\bibitem[{Rabuge and Louren{\c{c}}o(2024)}]{rabuge2024comparison}
Rabuge, M., Louren{\c{c}}o, N.: A comparison of feature engineering techniques for hearing loss. In: Proceedings of the Genetic and Evolutionary Computation Conference, pp. 0--0 (2024)

\bibitem[{Ryan et~al.(1998)Ryan, Collins, and Neill}]{ryan1998grammatical}
Ryan, C., Collins, J.J., Neill, M.O.: Grammatical evolution: Evolving programs for an arbitrary language. In: Genetic Programming: First European Workshop, EuroGP’98 Paris, France, April 14--15, 1998 Proceedings 1, pp. 83--96, Springer (1998)

\bibitem[{\text{World health Organization - Deafness and hearing loss}(2024)}]{whohl}
\text{World health Organization - Deafness and hearing loss}: \url{https://www.who.int/news-room/fact-sheets/detail/deafness-and-hearing-loss} (2024)

\bibitem[{Tomiazzi et~al.(2019)Tomiazzi, Pereira, Judai, Antunes, and Favareto}]{tomiazzi2019performance}
Tomiazzi, J.S., Pereira, D.R., Judai, M.A., Antunes, P.A., Favareto, A.P.A.: Performance of machine-learning algorithms to pattern recognition and classification of hearing impairment in brazilian farmers exposed to pesticide and/or cigarette smoke. Environmental Science and Pollution Research \textbf{26}, 6481--6491 (2019)

\bibitem[{Tran et~al.(2016)Tran, Zhang, and Xue}]{tran2016multiple}
Tran, B., Zhang, M., Xue, B.: Multiple feature construction in classification on high-dimensional data using gp. In: 2016 IEEE symposium series on computational intelligence (SSCI), pp. 1--8, IEEE (2016)

\end{thebibliography}

\end{document}